\documentclass[letterpaper]{article}
\usepackage[preprint]{aaai2027}
\usepackage[hyphens]{url}
\usepackage{graphicx}
\usepackage{natbib}
\usepackage{caption}
\usepackage{booktabs}
\usepackage{multirow}

\usepackage{amsmath}
\usepackage{amssymb}
\usepackage{tabularx}

\title{SimPOTrans: Self-Adaptation for Offline-to-Simultaneous Speech Translation \\via Reinforcement Learning}
\author{
    Haotian Tan,
    Hiroki Ouchi,
    Sakriani Sakti\corresponding
}
\affiliations{
    Nara Institute of Science and Technology, Japan\\
    tan.haotian.tf5@naist.ac.jp, hiroki.ouchi@is.naist.jp, ssakti@is.naist.jp
}

\begin{document}

\maketitle

\begingroup
\renewcommand{\thefootnote}{}
\footnotetext{Part of this work was supported by JSPS KAKENHI Grant Numbers JP21H05054, JP23K21681, JP25H01139, and JP26H02537, as well as JST NEXUS (JPMJNX25C1).}
\endgroup

\begin{abstract}
Simultaneous speech translation (SimulST) aims to translate ongoing speech with low latency while preserving translation quality. 
Recent studies have adopted large language models (LLMs) for SimulST and formulated it as a multi-turn dialogue task for efficient generation, achieving state-of-the-art performance. 
However, existing dialogue-style SimulST approaches typically rely on specialized training schemes to acquire simultaneous translation behavior, either using synthesized read/write trajectories, which may provide suboptimal training signals, or jointly optimizing translation quality and latency, whose competing objectives may cause optimization instability.
In this paper, we introduce SimPOTrans, a reinforcement learning (RL) framework that enables offline speech translation models to self-adapt to dialogue-style SimulST.
Through a simple two-stage RL post-training pipeline, SimPOTrans rewards an offline model for exhibiting translation behaviors beneficial to simultaneous scenarios when given partial speech inputs, thereby enabling offline-to-simultaneous self-adaptation without synthesized read/write trajectories or explicit latency optimization.
At inference time, SimPOTrans requires no external decision policy and enables flexible quality--latency control by simply adjusting the chunk size of speech input.
Experiments on the ACL 60/60 and MCIF benchmarks across three language directions (En$\rightarrow$De, En$\rightarrow$Zh, and En$\rightarrow$Ja) demonstrate that SimPOTrans outperforms several strong baselines, achieving a superior quality--latency trade-off.

\par\medskip
\noindent
\newcommand{\resourceurl}[1]{%
    \pdfstartlink attr{/Border [0 0 0]} user{/Subtype /Link /A << /S /URI /URI (#1) >>}\url{#1}\pdfendlink}
\begin{tabular}{@{}l@{\hspace{0.5em}}l@{}}
    \textbf{Code:} & \resourceurl{https://github.com/Tan-ht/SimPOTrans} \\
    \textbf{Demo:} & \resourceurl{https://tan-ht.github.io/naist-simulst-demo/}
\end{tabular}
\end{abstract}

\section{Introduction}
\label{sec:introduction}
Simultaneous speech translation (SimulST) aims to mimic human interpreters by translating ongoing speech in real time.
Unlike offline speech translation (ST), which processes complete utterances, SimulST generates translations from partial speech input while balancing translation quality and latency with a read/write decision policy.

Earlier SimulST approaches rely on predefined decision policies, either learned through dedicated training strategies~\citep{ma2020simulmt,zeng2021realtrans} or applied to offline-trained ST models at inference time~\citep{wait-k, localagreement,edatt,papi2023alignatt}. They often require training multiple models for different latency regimes or depend heavily on handcrafted rules.
Recent studies have shown that LLM-based translation backbones can substantially improve the quality-latency trade-off for SimulST~\citep{fasst,tan-etal-2025-naist,transllama,sasst,simulsense}.
In particular, LLM-based SimulST has been formulated as a multi-turn dialogue, with partial speech inputs as user prompts and incremental translations as assistant responses~\citep{wang-etal-2025-conversational,fu2026efficient,ouyang-etal-2025-infinisst}.
This paradigm enables efficient generation through KV-cache reuse and allows models to learn read/write policies from synthesized trajectories.
However, these methods rely on either potentially suboptimal synthesized trajectories or explicit joint optimization of quality and latency, whose competing objectives may cause optimization instability.

In this paper, we introduce \textbf{SimPOTrans}, a reinforcement learning (RL) framework that enables offline speech translation models to self-adapt to dialogue-style SimulST through Group Relative Policy Optimization (GRPO)~\citep{grpo}.
Our intuition is that an offline model may occasionally exhibit translation behaviors well suited to simultaneous scenarios when given partial speech inputs.
By rewarding these spontaneously exhibited simultaneous behaviors, SimPOTrans guides the offline model to self-adapt to SimulST without requiring synthesized read/write trajectories or explicit latency optimization.
Specifically, SimPOTrans applies a simple two-stage RL post-training pipeline to an offline ST model using standard offline speech translation data. 
The first stage strengthens the model's overall translation capability on complete utterances. 
Building on this foundation, the second stage exposes the model to partial speech inputs in a multi-turn setting, enabling it to self-adapt from offline to simultaneous translation.
By learning from its self-generated dialogue-style trajectories, the model learns what content to translate and how to formulate it at each turn.
Consequently, it acquires strong simultaneous translation capability without relying on synthesized read/write trajectories or explicit latency optimization.
At inference time, SimPOTrans requires no external decision policy, and its quality-latency trade-off can be controlled simply by varying the chunk size of speech input.
Our main contributions are summarized as follows:
\begin{itemize}
    \item We introduce \textbf{SimPOTrans}, an RL framework that enables offline ST models to self-adapt to SimulST by discovering and reinforcing effective simultaneous translation behavior through self-generated trajectories.

    \item We develop a simple two-stage RL post-training pipeline. It first strengthens the overall translation capability and then exposes the offline ST model to partial speech inputs, allowing the model to learn what content to translate and how to translate at each turn without using synthesized data or an explicit latency objective.

    \item We show that SimPOTrans not only improves translation quality but also induces translation behavior better suited to simultaneous scenarios.
\end{itemize}

\section{Related Work}
\label{sec:related-work}

\subsection{Dialogue-Style SimulST}
\label{subsec:dialogue-simulst}
Dialogue-style LLM-based SimulST formulates simultaneous translation as a multi-turn dialogue task, where speech segments and incremental translations are interleaved as user prompts and assistant responses, respectively. 
Existing approaches typically require synthesized read/write trajectories for training. 
A read/write trajectory records the interleaved process as $\tau=(\mathbf{S}_1,\mathbf{Y}_1,\ldots,\mathbf{S}_M,\mathbf{Y}_M)$, with each $\mathbf{S}_m$ denoting a speech segment read by the model and $\mathbf{Y}_m$ denoting the incremental translation written in response.
InfiniSST~\citep{ouyang-etal-2025-infinisst} constructs such data from word alignments over offline parallel data.
EASiST~\citep{fu2026efficient} and CLASI~\citep{CLASI} instead prompt LLMs to synthesize the read/write trajectory.
This paradigm enables efficient inference over unbounded speech input by reusing the key-value (KV) cache of LLMs across turns and has achieved state-of-the-art performance.
However, trajectories synthesized using automatic alignment tools or LLMs may contain segmentation and alignment errors, resulting in suboptimal training signals.
In contrast, SimPOTrans enables dialogue-style SimulST using only conventional offline ST data, without any synthesized read/write trajectories.

\subsection{Reinforcement Learning for SimulST}
\label{subsec:rl_for_simulst}
Existing studies of reinforcement learning (RL) for machine translation have primarily focused on text translation~\citep{he2024improving,ive-etal-2021-exploiting, xu-etal-2025-seqpo,yu2025simulpl}, while only a few studies have applied RL to SimulST.
CLASI~\citep{CLASI} conducts RL post-training on a supervised SimulST model using high-quality human interpretation data and a combination of process and outcome rewards.
However, it suffers from optimization instability and reward hacking due to the difficulty of balancing these rewards.
Hibiki-Zero~\citep{hibiki-zero} applies RL with intermediate BLEU rewards computed against time-indexed reference prefixes to optimize for the quality--latency trade-off.
However, it requires temporal supervision constructed from coarsely synchronized target speech, source-side timestamps, and partial references.
HPO~\citep{hpo} follows InfiniSST~\citep{ouyang-etal-2025-infinisst} to build a dialogue-style SimulST backbone using synthesized training trajectories and then applies RL to jointly optimize translation quality and latency.
Existing methods either require synthesized data, temporal supervision, or an explicit latency optimization objective.
In contrast, SimPOTrans enables an offline ST model to self-adapt to dialogue-style SimulST without requiring synthesized training data, temporal supervision, or explicit latency optimization.

\subsection{Offline Models for SimulST}
\label{subsec:offline-to-simulst}
Earlier studies primarily utilized offline-trained models for SimulST through inference-time decision policies. MoSST~\citep{dong-etal-2022-learning} applies CIF~\citep{cif} to identify word boundaries and enables SimulST using a test-time wait-$k$ policy~\citep{wait-k}. Local Agreement (LA)~\citep{localagreement,cunikit} commits only the hypotheses shared across two consecutive chunks, while EDAtt~\citep{edatt} and AlignAtt~\citep{papi2023alignatt} use cross-attention information to discard unstable trailing outputs. These approaches allow a single offline model to operate under different latency requirements without adapting the model itself. However, their performance depends heavily on the design of the inference-time policy, while the model is not explicitly trained to handle partial speech inputs. In contrast, SimPOTrans adapts the model to learn what and how to translate from partial speech inputs. It performs SimulST through simple chunk-by-chunk generation and supports different latency regimes by varying the input chunk size.

\begin{figure*}[t]
\centering
\includegraphics[width=\linewidth]{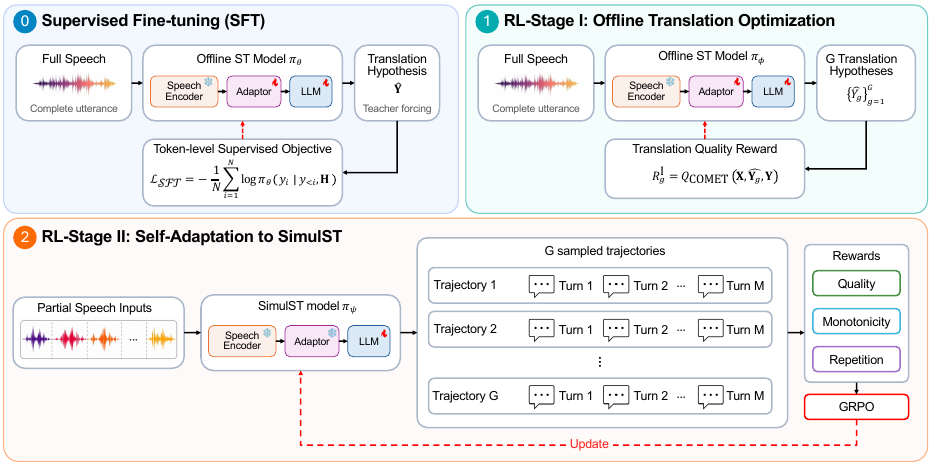}
\caption{Overview of the SimPOTrans training pipeline.
\textbf{(1) Stage 0: Offline Translation Pretraining (SFT).}
SFT establishes the offline translation capability of the LLM-based speech translation model.
\textbf{(2) Stage 1: Offline Translation Optimization (RL).}
RL-Stage I uses GRPO with a translation-quality reward to further enhance offline translation quality on complete utterances.
\textbf{(3) Stage 2: Offline-to-Simultaneous Translation Adaptation (RL).}
RL-Stage II exposes the offline-trained model to partial speech inputs in a multi-turn setting and optimizes it with quality and monotonicity rewards together with a repetition penalty, enabling the model to self-adapt to dialogue-style SimulST.
}
\label{fig:method-fig}
\end{figure*}

\section{Proposed Method}
\label{sec:method}
As illustrated in Figure~\ref{fig:method-fig}, SimPOTrans starts from a conventional offline speech translation model and adapts it to dialogue-style SimulST through a two-stage RL-based policy optimization procedure.
We first perform supervised fine-tuning (SFT) to establish offline translation capability.
RL-Stage I then improves sequence-level translation quality on complete utterances, while RL-Stage II exposes the model to partial speech inputs in a multi-turn setting and reinforces self-generated trajectories that are better suited to SimulST.
We train SimPOTrans on a standard offline speech translation dataset $\mathcal{D}$.
Each training example is a triplet $(\mathbf{S},\mathbf{X},\mathbf{Y})\in\mathcal{D}$, where $\mathbf{S}=(s_1,\ldots,s_T)$ is the source speech, $\mathbf{X}=(x_1,\ldots,x_L)$ is its source transcript, and $\mathbf{Y}=(y_1,\ldots,y_N)$ is the reference translation.

\subsection{Supervised Fine-Tuning for Offline Translation}
\label{subsec:sft}
Following recent LLM-based speech translation systems~\citep{tan-etal-2025-naist, ouyang-etal-2025-infinisst}, we adopt a translation backbone $\pi_\theta$ consisting of a speech encoder, an adaptor, and an LLM.
Given the source speech $\mathbf{S}$, the speech encoder $f_{\mathrm{enc}}$ extracts acoustic representations, which are projected by the adaptor $f_{\mathrm{ada}}$ into the embedding space of the LLM:
\begin{equation}
    \mathbf{H}
    =
    f_{\mathrm{ada}}
    \left(
        f_{\mathrm{enc}}(\mathbf{S})
    \right).
    \label{eq:speech-representation}
\end{equation}
The resulting speech representations are incorporated into the input prompt, and the LLM autoregressively generates a translation hypothesis
$\hat{\mathbf{Y}}=(\hat{y}_1,\ldots,\hat{y}_{\hat{N}})$, where
\begin{equation}
    \hat{y}_i
    \sim
    \pi_\theta
    \left(
        \cdot
        \mid
        \hat{y}_{<i},\mathbf{H}
    \right).
    \label{eq:sft-generation}
\end{equation}
The model is optimized using the standard cross-entropy loss:
\begin{equation}
    \mathcal{L}_{\mathrm{SFT}}
    =
    -\frac{1}{N}
    \sum_{i=1}^{N}
    \log \pi_\theta
    \left(
        y_i
        \mid
        y_{<i},\mathbf{H}
    \right).
    \label{eq:sft-loss}
\end{equation}
During SFT, the speech encoder remains frozen, while the adaptor is fully fine-tuned and the LLM is adapted using LoRA.
The resulting model $\pi_\theta$ provides the initialization for the subsequent RL post-training.

\subsection{Two-Stage Policy Optimization}
\label{subsec:rl-post-training}
Both RL stages employ GRPO for optimization.
For each training example, the old policy $\pi_{\omega_{\mathrm{old}}}$ samples a group of $G$ outputs $\{\mathcal{O}_g\}_{g=1}^{G}$, where $\omega$ denotes the trainable parameters of the current stage.
Given the stage-specific reward $R_g$ for each output, its group-relative advantage is computed as
\begin{equation}
    A_g
    =
    \frac{
        R_g-\operatorname{mean}(\{R_j\}_{j=1}^{G})
    }{
        \operatorname{std}(\{R_j\}_{j=1}^{G})+\epsilon
    },
    \label{eq:group-relative-advantage}
\end{equation}
where $\epsilon$ is a small constant for numerical stability.
This sequence-level advantage is shared across all tokens in the corresponding output.
Let
$\operatorname{clip}_{\varepsilon}(x)
=\operatorname{clip}(x,1-\varepsilon,1+\varepsilon)$,
where $\varepsilon$ controls the clipping range.
For the $t$-th token of $\mathcal{O}_g$, the per-token clipped surrogate term is defined as
\begin{equation}
    \ell_{g,t}(\omega)
    =
    \min\left\{
        \frac{\pi_\omega}{\pi_{\omega_{\mathrm{old}}}}A_g,\,
        \operatorname{clip}_{\varepsilon}\left(
            \frac{\pi_\omega}{\pi_{\omega_{\mathrm{old}}}}
        \right)A_g
    \right\}.
    \label{eq:grpo-surrogate}
\end{equation}
where $\pi_\omega/\pi_{\omega_{\mathrm{old}}}$ denotes the token-level importance sampling ratio.
The GRPO training loss is then defined as:
\begin{equation}
    \mathcal{L}_{\mathrm{GRPO}}(\omega)
    =
    -\frac{1}{G}
    \sum_{g=1}^{G}
    \frac{1}{L_g}
    \sum_{t=1}^{L_g}
    \ell_{g,t}(\omega).
    \label{eq:grpo-objective}
\end{equation}
where $L_g$ is the number of generated tokens in $\mathcal{O}_g$.

\subsubsection{Stage I: Offline Translation Optimization}
\label{subsec:offline-rl}
We further optimize the offline translation capability before adapting it to simultaneous translation.
We initialize $\pi_\phi$ from the SFT model $\pi_\theta$, i.e., $\phi\leftarrow\theta$.
Given a complete source utterance $\mathbf{S}$ and its speech representations $\mathbf{H}$, the old policy $\pi_{\phi_{\mathrm{old}}}$ samples $G$ translation hypotheses:
\begin{equation}
    \hat{\mathbf{Y}}_g
    \sim
    \pi_{\phi_{\mathrm{old}}}
    \left(
        \cdot
        \mid
        \mathbf{H}
    \right),
    \qquad
    g=1,\ldots,G.
    \label{eq:stage-one-sampling}
\end{equation}
We use COMET~\citep{rei2020comet} as the translation quality reward:
\begin{equation}
R^{\mathrm{I}}_g
=
Q_{\mathrm{COMET}}
\left(
\mathbf{X},
\hat{\mathbf{Y}}_g,
\mathbf{Y}
\right).
\end{equation}
Using $\{R^{\mathrm{I}}_g\}_{g=1}^{G}$, we compute the group-relative advantages according to Eq.~\eqref{eq:group-relative-advantage} and optimize $\pi_\phi$ using Eq.~\eqref{eq:grpo-objective}.
During RL-Stage I, the speech encoder remains frozen, while the adaptor and the LoRA parameters of the LLM are updated.
The resulting model $\pi_\phi$ provides the initialization for the subsequent RL-Stage II.

\subsubsection{Stage II: Self-Adaptation to SimulST}
\label{subsec:simul-rl}

In Stage II, the offline-optimized model self-adapts to SimulST by processing partial speech inputs and learning from its self-generated multi-turn translation trajectories.
Specifically, we initialize the SimulST model $\pi_\psi$ from $\pi_\phi$, i.e., $\psi\leftarrow\phi$, and require the old policy $\pi_{\psi_{\mathrm{old}}}$ to operate in a multi-turn dialogue manner.
For each source utterance $\mathbf{S}$, we sample $G$ translation trajectories. 
We first divide $\mathbf{S}$ into $M$ fixed-duration chunks $\mathbf{S}_1,\ldots,\mathbf{S}_M$.
At turn $m$ of the multi-turn generation process, the speech encoder processes all currently available speech, $\mathbf{S}_{\leq m}$. From the resulting representations, we retain only those corresponding to the latest chunk $\mathbf{S}_m$ and pass them through the adaptor:
\begin{equation}
    \mathbf{H}_m
    =
    f_{\mathrm{ada}}
    \left(
        \operatorname{Slice}_m
        \left[
            f_{\mathrm{enc}}
            \left(
                \mathbf{S}_{\leq m}
            \right)
        \right]
    \right),
    \label{eq:incremental-encoding}
\end{equation}
where $\operatorname{Slice}_m[\cdot]$ selects the encoder frames corresponding to $\mathbf{S}_m$.
For the $g$-th trajectory, the LLM generates a translation segment at each turn conditioned on the interleaved speech representations and translation segments from the preceding turns:
\begin{equation}
    \hat{\mathbf{Y}}_{g,m}
    \sim
    \pi_{\psi_{\mathrm{old}}}
    \left(
        \cdot
        \mid
        \mathbf{H}_1,\hat{\mathbf{Y}}_{g,1},
        \ldots,
        \mathbf{H}_{m-1},\hat{\mathbf{Y}}_{g,m-1},
        \mathbf{H}_m
    \right).
    \label{eq:turn-generation}
\end{equation}
After all $M$ turns, the complete translation is obtained by concatenating all generated segments:
\begin{equation}
    \hat{\mathbf{Y}}_g
    =
    \operatorname{Concat}
    \left(
        \hat{\mathbf{Y}}_{g,1},
        \ldots,
        \hat{\mathbf{Y}}_{g,M}
    \right).
    \label{eq:complete-translation}
\end{equation}

\paragraph{Translation quality reward.}
We evaluate translation quality using the same COMET reward as in RL-Stage I:
\begin{equation}
    Q_g
    =
    Q_{\mathrm{COMET}}
    \left(
        \mathbf{X},
        \hat{\mathbf{Y}}_g,
        \mathbf{Y}
    \right).
    \label{eq:stage2-quality-reward}
\end{equation}
The reward is computed on the complete translation rather than individual output segments.
Therefore, no chunk-level references are required.

\paragraph{Monotonicity reward.}
Prior studies have shown that improving translation monotonicity correlates with improvements in both quality and latency in SimulST~\citep{doi2024word,makinae2024simul}.
We therefore encourage the generated translation to follow an ordering consistent with the source.
Following prior work~\citep{makinae2024simul}, we obtain word alignments between the source transcript $\mathbf{X}$ and the generated translation $\hat{\mathbf{Y}}_g$ using Awesome-Align~\citep{awesome-align}.
Let $\mathbf{a}_g$ and $\mathbf{b}_g$ denote the source and target positions of the aligned word pairs, respectively.
We compute the monotonicity score using Spearman's rank correlation coefficient and map it to the same $[0,1]$ scale as the quality reward:
\begin{equation}
    \mathcal{M}_g
    =
    \frac{
        \rho_{\mathrm{sp}}
        \left(
            \mathbf{a}_g,\mathbf{b}_g
        \right)
        +1
    }{2}.
    \label{eq:monotonicity-score}
\end{equation}
Only trajectories with a monotonicity score lower than
$\kappa_{\mathrm{mono}}=0.9$ are penalized:
\begin{equation}
    R^{\mathrm{mono}}_g
    =
    -\max
    \left(
        0,\,
        \kappa_{\mathrm{mono}}-\mathcal{M}_g
    \right).
    \label{eq:monotonicity-reward}
\end{equation}

\paragraph{Repetition penalty.}
We empirically observe that repeating content from the preceding turn is the most frequent error during simultaneous adaptation.
To suppress this behavior, we measure the $n$-gram overlap between each pair of adjacent output segments:
\begin{equation}
    o_{g,m}
    =
    \frac{
        \left|
            \mathcal{N}_n(\hat{\mathbf{Y}}_{g,m})
            \cap
            \mathcal{N}_n(\hat{\mathbf{Y}}_{g,m-1})
        \right|
    }{
        \left|
            \mathcal{N}_n(\hat{\mathbf{Y}}_{g,m})
        \right|
    },
    \label{eq:repetition-overlap}
\end{equation}
where $\mathcal{N}_n(\cdot)$ denotes the set of $n$-grams extracted from an output segment.
We set $o_{g,m}=0$ if either segment contains no $n$-grams.
We compute the repetition penalty by averaging the excess overlap above a predefined threshold across all adjacent output pairs:
\begin{equation}
    P^{\mathrm{rep}}_g
    =
    \frac{1}{M-1}
    \sum_{m=2}^{M}
    \max
    \left(
        0,\,
        o_{g,m}-\kappa_{\mathrm{rep}}
    \right),
    \label{eq:repetition-penalty}
\end{equation}
where $\kappa_{\mathrm{rep}}=0.65$ during training.

\paragraph{Overall training objective.}
The overall trajectory-level reward is defined as
\begin{equation}
    R^{\mathrm{II}}_g
    =
    Q_g
    +
    \lambda_{\mathrm{mono}}R^{\mathrm{mono}}_g
    -
    \lambda_{\mathrm{rep}}P^{\mathrm{rep}}_g,
    \label{eq:stage-two-reward}
\end{equation}
where $\lambda_{\mathrm{mono}}$ and $\lambda_{\mathrm{rep}}$ control the contributions of the monotonicity reward and repetition penalty, respectively.
This objective promotes translation quality and monotonicity while discouraging excessive repetition between adjacent turns.
Using $\{R^{\mathrm{II}}_g\}_{g=1}^{G}$, we compute the group-relative advantages according to Eq.~\eqref{eq:group-relative-advantage} and optimize $\pi_\psi$ using Eq.~\eqref{eq:grpo-objective}.
During RL-Stage II, both the speech encoder and the adaptor remain frozen, and only the LoRA parameters of the LLM are updated.

\subsection{SimulST Inference for Unbounded Speech}
\label{subsec:simulst-inference}
At inference time, the incoming speech is divided into fixed-duration chunks of $\delta$ seconds and translated in a dialogue-style manner, with each chunk initiating a new generation turn.
The quality--latency trade-off can be controlled by varying $\delta$.
To prevent the dialogue context from becoming excessively long, we periodically reset it based on the duration of the retained speech.
Once the duration exceeds $20$ seconds, the context is reset when the latest translation ends with sentence-ending punctuation.
If adding the next chunk would extend the retained speech beyond $30$ seconds before such punctuation is generated, the context is forcibly reset before processing that chunk.

\section{Experimental Setup}
\label{sec:experimental-setup}

\subsection{Data}
\label{subsec:data}
We train our models on the widely used CoVoST~2~\citep{covost2} corpus. We consider three English-to-X directions: En$\rightarrow$De, En$\rightarrow$Zh, and En$\rightarrow$Ja. Each contains approximately 430 hours of training speech.
For evaluation, we use two long-form speech translation benchmarks derived from scientific presentations. ACL 60/60~\citep{salesky-etal-2023-evaluating} consists of recordings of ACL 2022 talks paired with translations into multiple languages. We additionally evaluate on the MCIF development set~\citep{papi2026mcif}, which contains long-form scientific talks and serves as the development data for the IWSLT 2026 simultaneous speech translation track~\citep{iwslt2026}.

\subsection{Model Configuration}
\label{subsec:model-config}
Our backbone comprises the Whisper-large-v3 encoder~\citep{whisper}, a two-layer MLP adaptor, and Qwen3-8B~\citep{qwen3}, adapted with LoRA~\citep{lora} of rank $16$ and scale $32$. We use AdamW with learning rates of $1\times10^{-4}$, $2\times10^{-6}$, and $5\times10^{-6}$ for SFT, RL-Stage I, and RL-Stage II, respectively. All stages use weight decay $0.01$ and $(\beta_1,\beta_2)=(0.9,0.95)$. In both RL stages, we sample with a temperature of $0.7$ and top-$p=0.95$ and reuse each rollout batch for $10$ epochs. We set the GRPO clipping parameter to $0.2$, clip gradients at norm $1.0$, and apply no KL regularization. Group sizes are $12$ and $8$ for RL-Stages I and II, respectively, with XCOMET-XL~\citep{guerreiro-etal-2024-xcomet} as the quality reward. RL-Stage II uses $2.0$-second chunks and $\lambda_{\mathrm{mono}}=0.2$, $\lambda_{\mathrm{rep}}=0.05$.

\subsection{Evaluation Metrics}
\label{subsec:evaluation-metrics}

\paragraph{Translation Quality}
We primarily evaluate translation quality using XCOMET-XL and MetricX-24~\citep{metricx}. XCOMET-XL is a learned reference-based metric with scores ranging from 0 to 1, where higher is better. Since XCOMET-XL is also used as the quality reward during reinforcement learning, relying exclusively on it for evaluation may overestimate improvements resulting from direct optimization toward the reward model. We therefore include MetricX-24 as an independent, complementary metric. MetricX-24 is a learned translation evaluation metric that predicts an error score between 0 and 25, where lower is better. We also report BLEU~\citep{papineni-etal-2002-bleu} to facilitate comparison with prior work.

\paragraph{Latency}
We report non-computation-aware (NCA) LongYAAL~\citep{polak-etal-2025-better}, which measures latency over continuous, unbounded speech while mitigating the segmentation bias of conventional average-lagging metrics.

\subsection{Baselines}
\label{subsec:baselines}
We compare SimPOTrans with three representative LLM-based SimulST systems:
\begin{itemize}
    \item \textbf{SimulSense}~\citep{simulsense} decouples read/write decisions from translation generation by using a separately trained decision module to trigger translation upon detecting a new interpreting unit, achieving a strong quality--latency trade-off.
    
    \item \textbf{InfiniSST}~\citep{ouyang-etal-2025-infinisst} formulates SimulST as a multi-turn dialogue task and trains the model using synthesized interleaved read/write trajectories. It achieved the best translation quality in the low-latency track of IWSLT 2025~\citep{ouyang-etal-2025-cmus,agostinelli2025findings}.
    
    \item \textbf{HPO}~\citep{hpo} applies RL post-training to InfiniSST using hierarchical rewards that jointly optimize translation quality and latency, achieving state-of-the-art performance.
\end{itemize}
These baselines represent three major directions in recent LLM-based SimulST: dedicated decision modules, dialogue-style trajectory supervision, and RL-based joint optimization of quality and latency. For controlled comparison, we reproduce all three systems using the same training data and LLM backbone as SimPOTrans.

We additionally compare SimPOTrans with low-latency results from four systems submitted to the IWSLT 2026 Simultaneous Speech Translation Track~\citep{iwslt2026}: NeMo~\citep{grigoryan2026nemo}, CUHKSZ~\citep{yang2026cuhksz}, PINCH-AST~\citep{bentes2026pinch}, and CUNI-POCKET~\citep{ortega2026pocket}. These results satisfy the official low-latency criterion of at most $2$ seconds and provide a reference for comparison with recent shared-task systems, although they do not constitute controlled comparisons.

\begin{figure*}[!t]
\centering
\begingroup
\setlength{\tabcolsep}{2pt}
\renewcommand{\arraystretch}{0.85}
\begin{tabular}{@{}c c c@{}}
\includegraphics[width=0.30\textwidth]{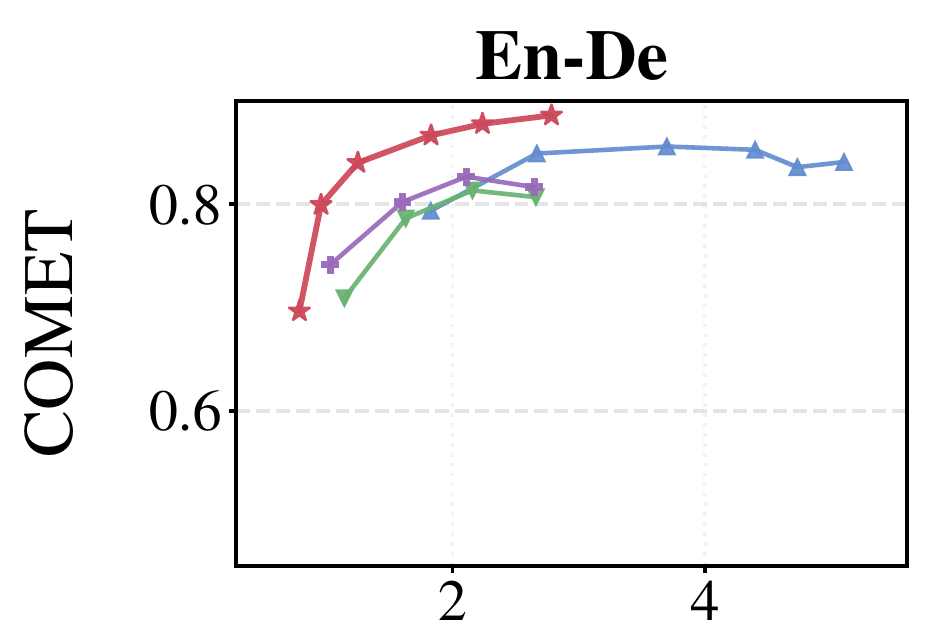} &
\includegraphics[width=0.30\textwidth]{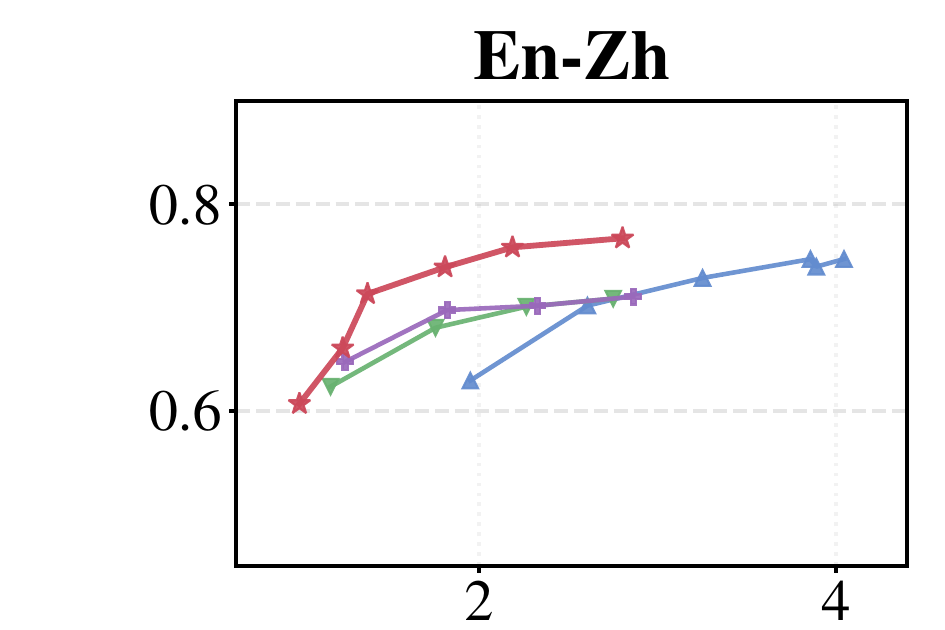} &
\includegraphics[width=0.30\textwidth]{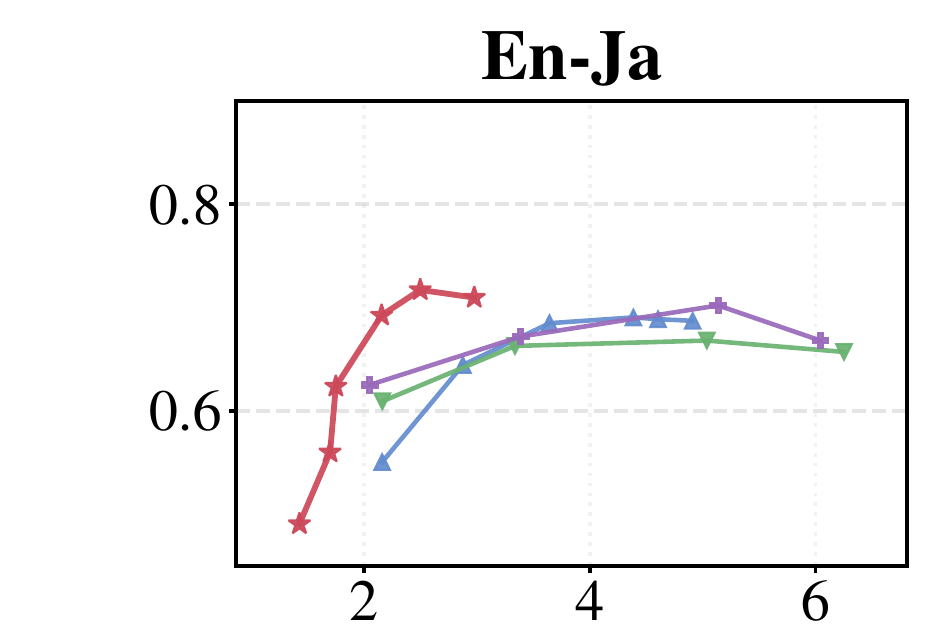} \\
\includegraphics[width=0.30\textwidth]{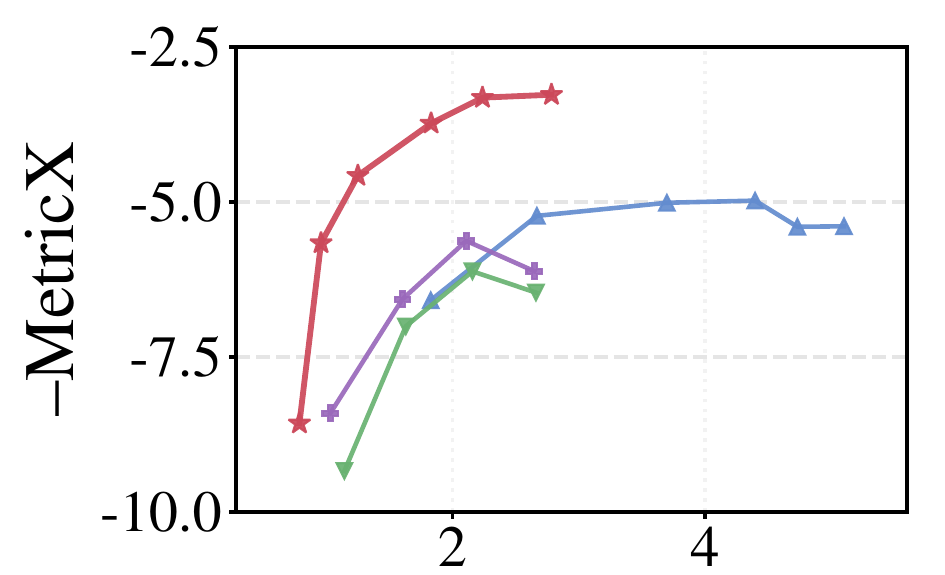} &
\includegraphics[width=0.30\textwidth]{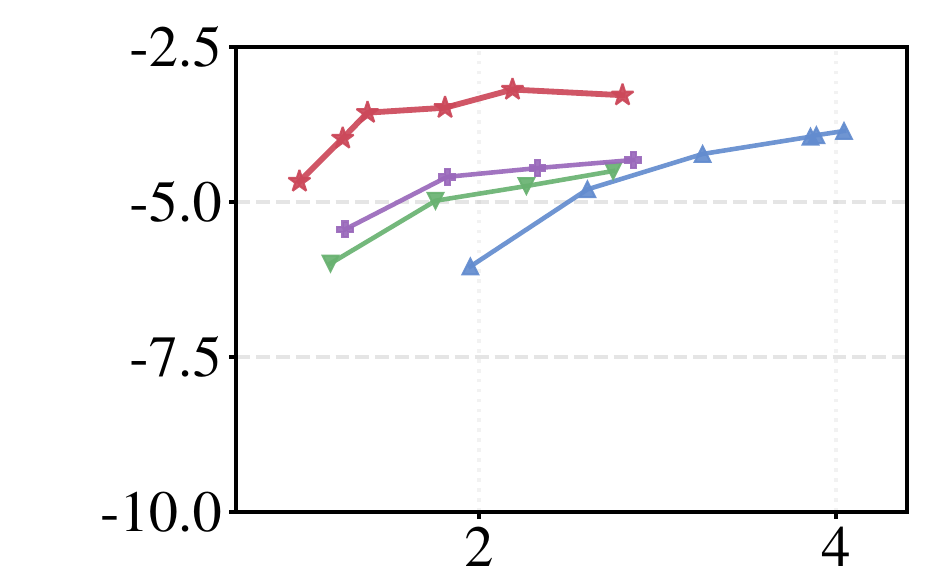} &
\includegraphics[width=0.30\textwidth]{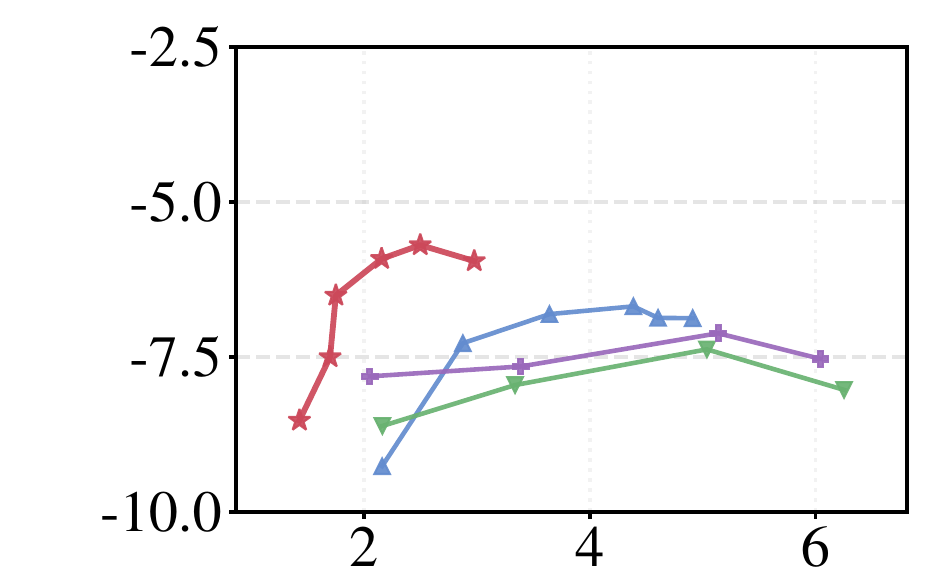} \\
\includegraphics[width=0.30\textwidth]{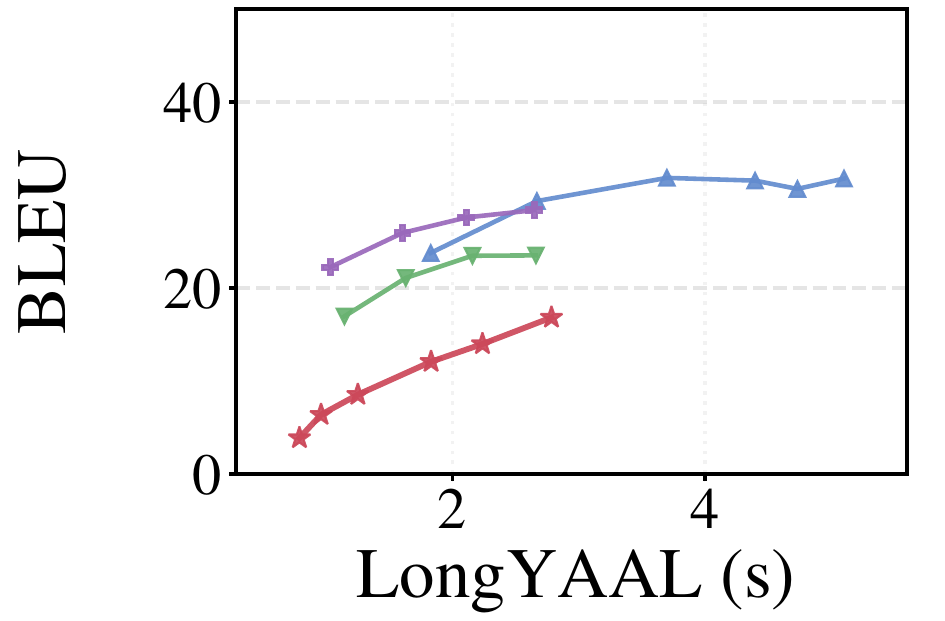} &
\includegraphics[width=0.30\textwidth]{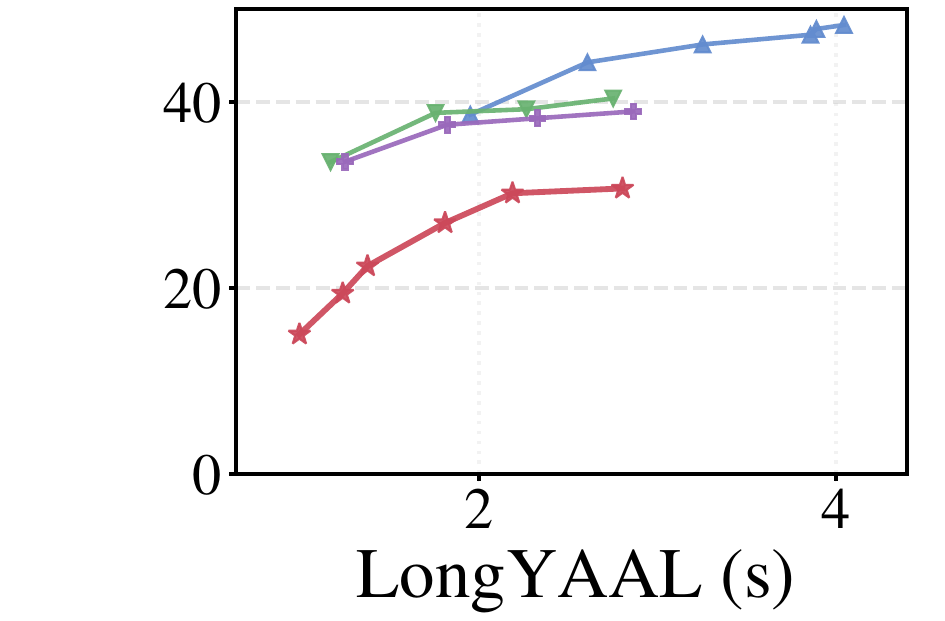} &
\includegraphics[width=0.30\textwidth]{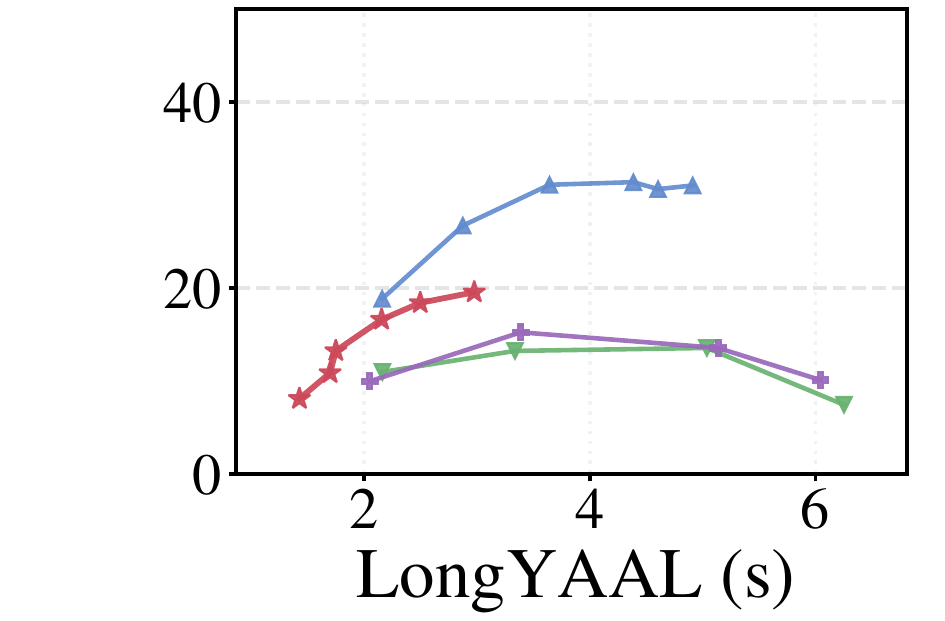} \\
\multicolumn{3}{c}{\includegraphics[width=0.65\textwidth]{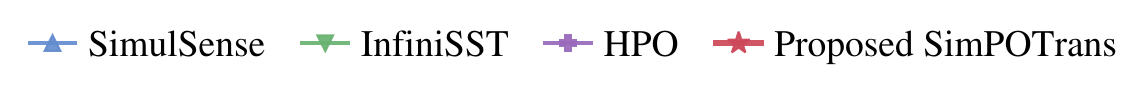}}
\end{tabular}
\endgroup
\caption{Comparison of quality--latency trade-offs on the ACL 60/60 benchmark for En$\rightarrow$De, En$\rightarrow$Zh, and En$\rightarrow$Ja. Rows report COMET, MetricX-24, and BLEU, respectively, with latency measured by LongYAAL. MetricX-24 scores are negated so that higher values consistently indicate better translation quality across all metrics. The SimPOTrans curves are obtained using chunk sizes of $\{1.0, 1.5, 2.0, 3.0, 4.0, 5.0\}$ seconds.}
\label{fig:main-quality-latency-grid}
\end{figure*}

\section{Main Results}
\label{sec:results}

\subsection{SimPOTrans Achieves Superior Quality--Latency Trade-offs}
\label{subsec:quality-latency}
Figure~\ref{fig:main-quality-latency-grid} compares the quality--latency trade-offs across three language directions and three quality metrics on ACL 60/60. SimPOTrans consistently outperforms all baselines in terms of both COMET and MetricX-24 across different latency regimes. BLEU exhibits a different trend, and we discuss this discrepancy in a later analysis.

\subsection{SimPOTrans Delivers Strong Low-Latency Performance without Additional Context}
\label{subsec:iwslt-comparison}
Table~\ref{tab:iwslt-low-latency-comparison} compares SimPOTrans with systems submitted to the IWSLT 2026 Simultaneous Speech Translation Track on the MCIF development set. We focus on the low-latency regime, in which all systems operate at a LongYAAL of at most two seconds. For En$\rightarrow$De, SimPOTrans achieves a COMET score of 0.87 with a LongYAAL of 1.6 seconds, outperforming all systems except NeMo, a cascaded system that incorporates additional context extracted from PDF documents. For En$\rightarrow$Zh, SimPOTrans achieves the highest COMET score of 0.76 while maintaining a LongYAAL of 1.7 seconds. These results demonstrate that SimPOTrans delivers strong low-latency translation performance without relying on any additional contextual information.

\begin{table}[!t]
\centering
\begingroup
\small
\setlength{\tabcolsep}{3.5pt}
\begin{tabular}{@{}llccc@{}}
\toprule
Lang. & System & COMET & LongYAAL & PDF \\
\midrule
\multirow{5}{*}{En$\rightarrow$De}
 & NeMo\textsuperscript{*} & 0.93 & 2.0 & \(\surd\) \\
 & CUHKSZ & 0.85 & 1.5 & \(\surd\) \\
 & PINCH-AST\textsuperscript{*} & 0.84 & 1.8 & \(\times\) \\
 & CUNI-POCKET & 0.77 & 1.7 & \(\times\) \\
 & Ours & 0.87 & 1.6 & \(\times\) \\
\midrule
\multirow{3}{*}{En$\rightarrow$Zh}
 & CUHKSZ & 0.74 & 1.5 & \(\surd\) \\
 & PINCH-AST\textsuperscript{*} & 0.72 & 1.8 & \(\times\) \\
 & Ours & 0.76 & 1.7 & \(\times\) \\
\bottomrule
\end{tabular}
\endgroup
\caption{Comparison with IWSLT 2026 Simultaneous Track systems on the MCIF development set in the low-latency regime \((\leq 2\,\mathrm{s})\). LongYAAL is reported in seconds. Asterisks indicate cascaded systems. PDF indicates whether PDF-derived extra context is used.}
\label{tab:iwslt-low-latency-comparison}
\end{table}

\begin{figure*}[t]
\centering
\begingroup
\setlength{\tabcolsep}{4pt}
\begin{tabular}{@{}ccc@{}}
\includegraphics[width=0.30\textwidth]{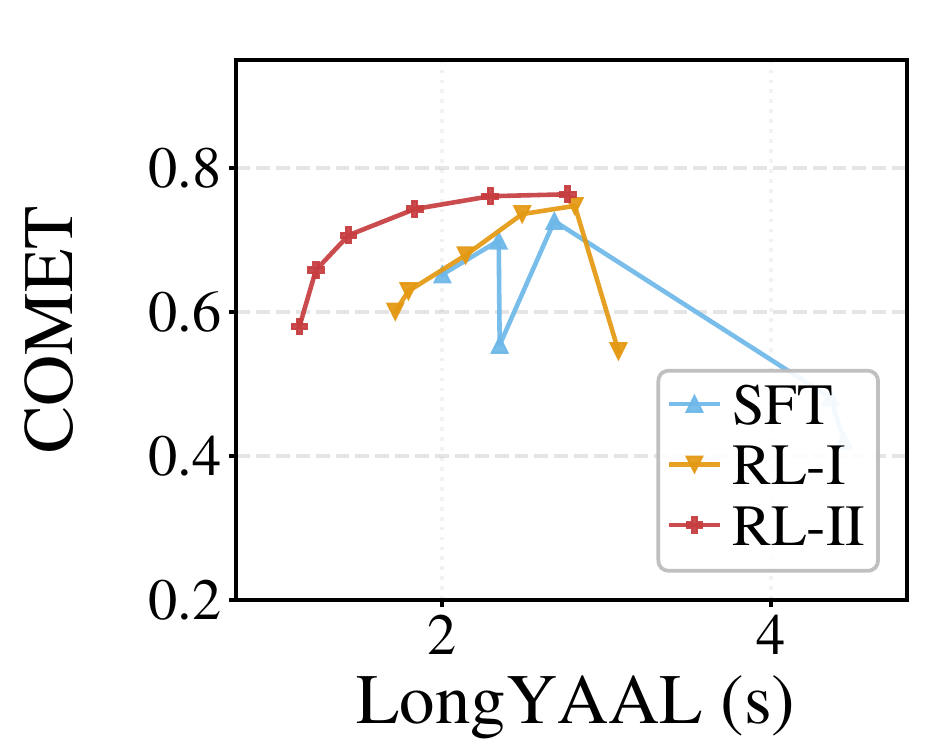} &
\includegraphics[width=0.30\textwidth]{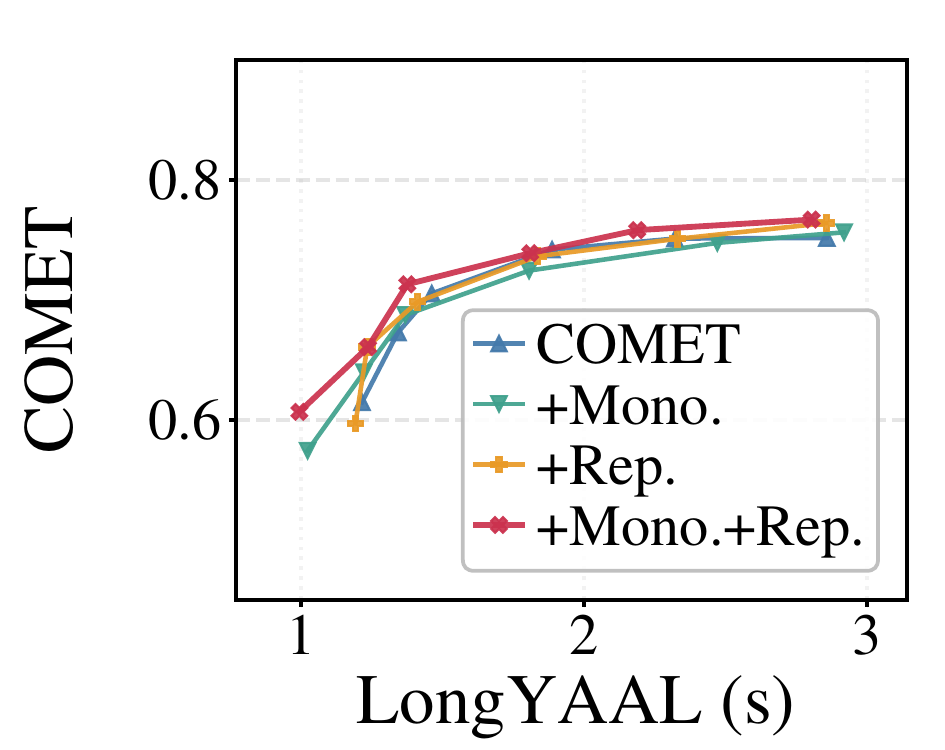} &
\includegraphics[width=0.30\textwidth]{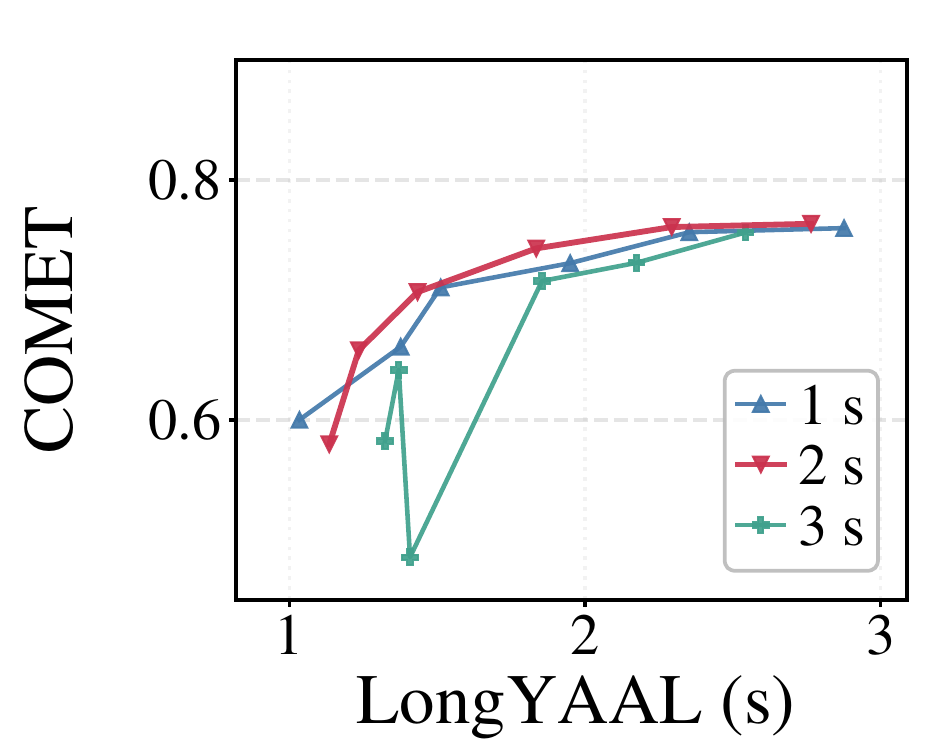} \\
\textbf{(a)} & \textbf{(b)} & \textbf{(c)}
\end{tabular}
\endgroup
\caption{Ablation studies. (a) Comparison of quality--latency trade-offs across different training stages. (b) Effects of the monotonicity reward (Mono.) and repetition penalty (Rep.) in RL-Stage II. (c) Effect of the speech chunk size in RL-Stage II.}
\label{fig:ablation-overview}
\end{figure*}

\begin{figure}[t]
\centering
\includegraphics[width=0.75\columnwidth]{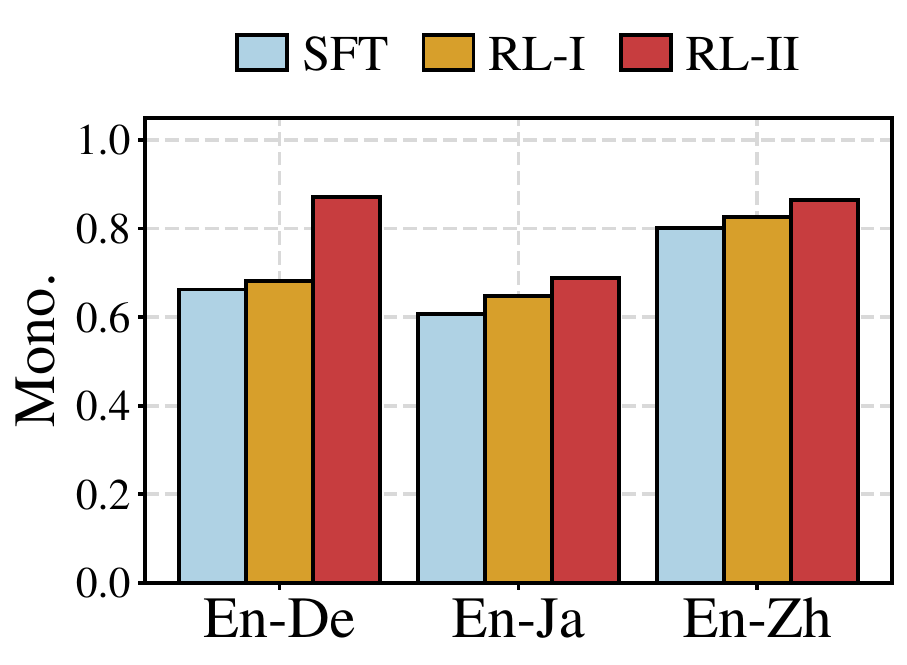}
\caption{Monotonicity across SimPOTrans training stages under simultaneous inference with a fixed chunk size of $1.5$ seconds.}
\label{fig:mono-compare}
\end{figure}

\begin{figure}[!b]
\centering
\includegraphics[width=\columnwidth]{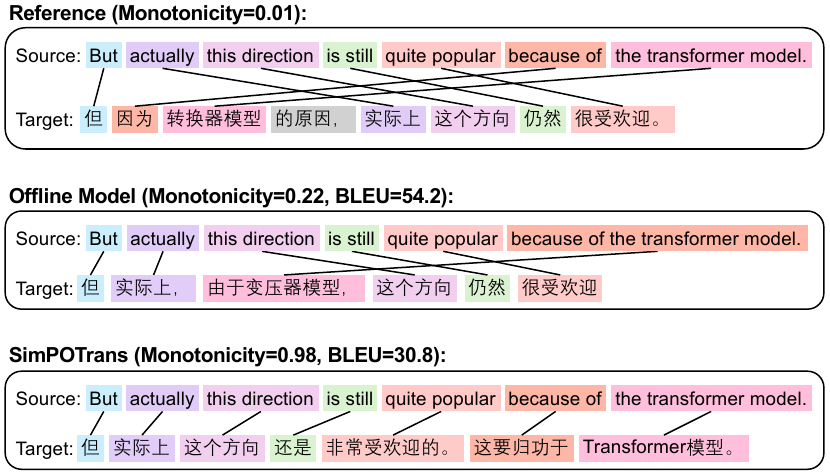}
\caption{An En$\rightarrow$Zh example comparing translation behaviors before and after offline-to-simultaneous self-adaptation. Colored spans and connecting lines indicate source--target alignments. The offline model largely follows the reference word order, which involves substantial reordering relative to the source. In contrast, SimPOTrans conveys the core meaning in an order closer to the source, producing a more monotonic translation better suited to simultaneous generation.}
\label{fig:mono-example}
\end{figure}

\subsection{SimPOTrans Induces Translation Behavior Better Suited to SimulST}
\label{subsec:translation-behavior}
Figure~\ref{fig:main-quality-latency-grid} reveals a discrepancy between semantic metrics and the $n$-gram-based lexical-overlap metric BLEU: SimPOTrans performs better on COMET and MetricX but substantially worse on BLEU.
A similar discrepancy is reported by HPO~\citep{hpo}, which attributes it to potential reward hacking.
Instead, our analysis suggests that the offline-to-simultaneous self-adaptation of SimPOTrans encourages the model to preserve the source meaning while expressing it in a more monotonic manner better suited to simultaneous translation. Such reformulation may deviate from the wording and word order of the reference, resulting in lower $n$-gram overlap and consequently lower BLEU scores.

Figure~\ref{fig:mono-compare} compares translation monotonicity across the three training stages of SimPOTrans in the simultaneous setting.
Monotonicity consistently increases across all three language directions, showing that the resulting translations more closely follow the source order.
The En$\rightarrow$Zh example in Figure~\ref{fig:mono-example} provides a concrete illustration.
The offline model produces a translation that closely follows the reference order and achieves a BLEU score of $54.2$, despite requiring substantial reordering relative to the source transcript.
In contrast, SimPOTrans conveys the core meaning in a more monotonic manner, but receives a substantially lower BLEU score of $30.8$.
This decrease reflects the reduced $n$-gram overlap caused by changes in wording and word order, illustrating how BLEU may penalize reformulations that are better suited to simultaneous translation.

\section{Ablation Study}
\label{sec:ablation}

\subsection{Effect of Simultaneous Adaptation}
\label{subsec:effect-simultaneous-adaptation}
Figure~\ref{fig:ablation-overview}(a) compares the quality--latency trade-offs across checkpoints after SFT, RL-Stage I, and RL-Stage II. After simultaneous adaptation in RL-Stage II, increasing the chunk size leads to consistent improvements in translation quality alongside increasing latency, forming a smooth quality--latency curve. This enables flexible control of the operating point simply by varying the chunk size. 
In contrast, the SFT and RL-Stage I models produce irregular curves and perform poorly on partial speech inputs. These results demonstrate that RL-Stage II is essential for adapting the offline model to simultaneous translation.

\subsection{Effect of Auxiliary Rewards}
\label{subsec:reward-ablation}
Figure~\ref{fig:ablation-overview}(b) evaluates the auxiliary terms in RL-Stage II. The monotonicity reward yields mixed quality gains, while the repetition penalty provides more consistent improvements. Combining both achieves the best overall quality--latency trade-offs, showing their complementary effects in promoting monotonic generation and reducing repetition.

\subsection{Effect of Chunk Size in RL-Stage II}
\label{subsec:chunk-size-ablation}
Figure~\ref{fig:ablation-overview}(c) compares the fixed chunk sizes used in RL-Stage II. Training with 2-second chunks achieves the best overall quality--latency trade-offs, while 1-second chunks provide insufficient context and 3-second chunks perform poorly at low latency. We therefore use a 2-second chunk size for RL-Stage II.

\section{Limitations}
\label{sec:limitations}
Although SimPOTrans demonstrates superior quality--latency trade-offs over several strong baselines, it currently processes speech using fixed-duration chunks, which may split semantically or acoustically coherent units. Future work could explore boundary-aware chunking strategies.
In addition, our evaluation focuses solely on speech-to-text translation. Future work could extend SimPOTrans to speech-to-speech translation.

\section{Conclusion}
\label{sec:conclusion}
In this work, we introduced SimPOTrans, a two-stage RL post-training framework that enables offline speech translation models to self-adapt to dialogue-style SimulST. SimPOTrans first improves translation quality on complete utterances and then learns to handle partial speech inputs through self-generated multi-turn trajectories, eliminating the need for synthesized data and explicit latency optimization. 
Experiments on ACL 60/60 and MCIF across three language directions demonstrate superior quality--latency trade-offs over strong LLM-based baselines. Further analyses show that SimPOTrans induces translation behavior better suited to simultaneous scenarios.


\bibliography{aaai2027}

\end{document}